%% file: main.tex
\documentclass[letterpaper, 10 pt, conference]{ieeeconf}
\usepackage[letterpaper, left=0.75in, right=0.75in, bottom=0.77in, top=0.78in]{geometry}
\usepackage{cite}
\IEEEoverridecommandlockouts                              

\usepackage{graphicx}
\usepackage{amsmath}
\usepackage{xcolor}
\usepackage{floatrow}
\usepackage{epstopdf}
\usepackage[colorlinks = true,
            linkcolor = black,
            urlcolor  = black,
            citecolor = black,
            anchorcolor = black]{hyperref}

\usepackage{algorithm,algpseudocode}
\usepackage[font=small]{caption}
\usepackage{accents}
\usepackage{amsfonts}
\usepackage{subcaption}
\usepackage{booktabs}
\usepackage{balance}
\usepackage{scalerel}
\usepackage{multirow}
\usepackage{url}
\usepackage{svg}
\usepackage{hhline}
\usepackage{hyperref}
\usepackage{soul}
\usepackage{verbatimbox}
\usepackage{makecell}
\usepackage{xcolor,colortbl}

\definecolor{LightCyan}{rgb}{0.88,1,1}
\definecolor{LightRed}{rgb}{1,0.88,0.88}
\newcolumntype{g}{>{\columncolor{LightCyan}}c}
\newcolumntype{r}{>{\columncolor{LightRed}}c} 

\newcommand{\vect}[1]{\boldsymbol{#1}}

\long\def\trecs{{\em D-EOF} }
\long\def\trecsend{{\em D-EOF}}

\long\def\eof{{\em EOF} }
\long\def\eofend{{\em EOF}}

\long\def\ignore#1{}
\begin{document}

\title{\LARGE \bf Tabletop Transparent Scene Reconstruction via Epipolar-Guided Optical Flow with Monocular Depth Completion Prior}
\author{
Xiaotong Chen$^{1}$,
Zheming Zhou$^{2}$, 
Zhuo Deng$^{2}$, 
Omid Ghasemalizadeh$^{2}$, \\
Min Sun$^{2}$, 
Cheng-Hao Kuo$^{2}$, 
Arnie Sen$^{2}$
\thanks{$^{1}$ X. Chen is with the Department of Robotics, University of Michigan, Ann Arbor, MI, USA. {\tt\footnotesize cxt@umich.edu}}
\thanks{$^{2}$ Z. Zhou, Z. Deng, O. Ghasemalizadeh, M. Sun, and C.H. Kuo are with Amazon Lab126, Sunnyvale, CA, USA. {\tt\footnotesize \{zhemiz, zhuod, ghasemal, minnsun, chkuo, senarnie\}@amazon.com}}
}


\maketitle
\input{0_abstract}


\section{Introduction}
\input{1_intro}

\section{Related Work}
\input{2_relatedwork}

\section{Transparent Scene Reconstruction Pipeline}
\input{3_method}

\section{Experiments}
\input{4_experiment}
\section{Conclusion}
\input{5_conclusion}

\balance


\bibliographystyle{IEEEtran}
\bibliography{ref}

\end{document}

%% file: 0_abstract.tex
\begin{abstract}

Reconstructing transparent objects using affordable RGB-D cameras is a persistent challenge in robotic perception due to inconsistent appearances across views in the RGB domain and inaccurate depth readings in each single-view. We introduce a two-stage pipeline for reconstructing transparent objects tailored for mobile platforms. In the first stage, off-the-shelf monocular object segmentation and depth completion networks are leveraged to predict the depth of transparent objects, furnishing single-view shape prior. Subsequently, we propose Epipolar-guided Optical Flow (EOF) to fuse several single-view shape priors from the first stage to a cross-view consistent 3D reconstruction given camera poses estimated from opaque part of the scene. Our key innovation lies in EOF which employs boundary-sensitive sampling and epipolar-line constraints into optical flow to accurately establish 2D correspondences across multiple views on transparent objects. Quantitative evaluations demonstrate that our pipeline significantly outperforms baseline methods in 3D reconstruction quality, paving the way for more adept robotic perception and interaction with transparent objects.


\end{abstract}

%% file: 1_intro.tex
\begin{figure}
    \centering
    \includegraphics[width=\columnwidth]{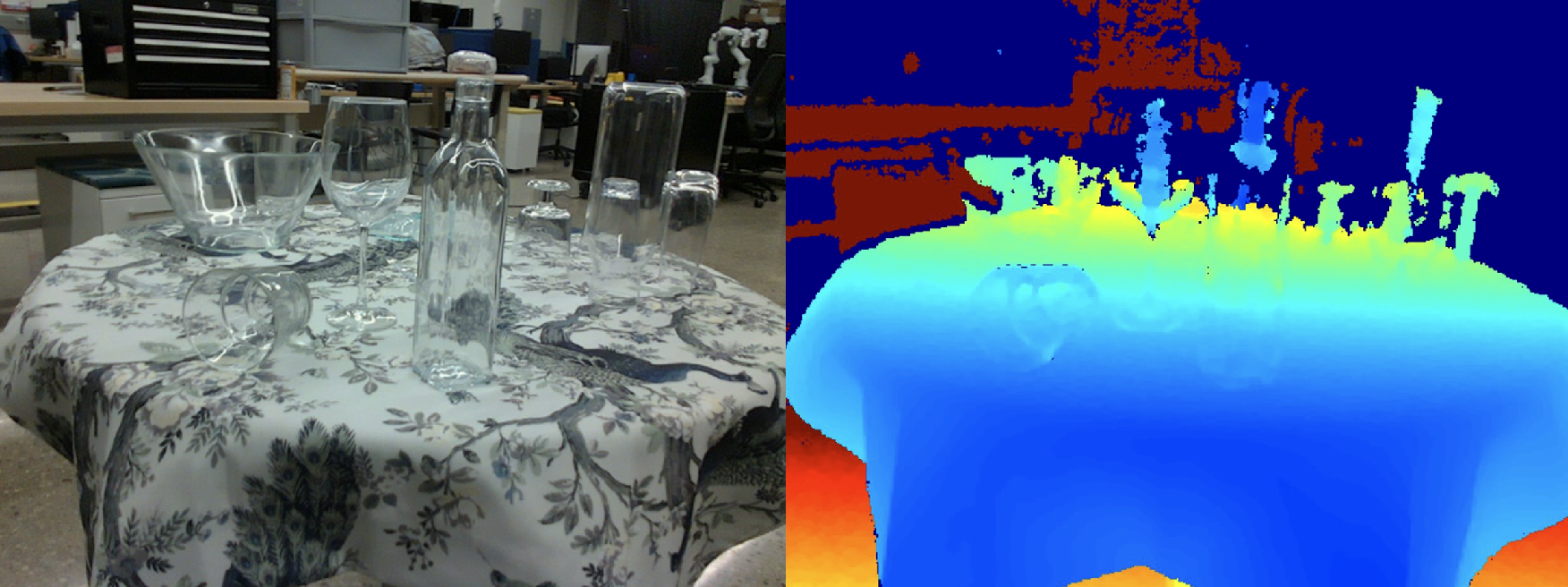}
    \caption{
    (Left) an RGB image of a transparent object scene. (Right) raw depth image from an Intel RealSense L515 camera. Compared to ground truth, most pixel values in the transparent region in the raw depth image are missing or inaccurate.
}
    \label{fig:trans_challenge}
\end{figure}

Transparent objects are prevalent in daily life, from small tabletop objects like cups, bottles, and bowls, to large glass windows and doors. They possess challenges to robotics visual perception mainly two-fold. In the RGB domain, transparent objects lack distinctive visual features, and their appearance is highly correlated to the background and environmental lighting, aggravating the problem. In the depth domain, the non-Lambertian surface cannot produce reliable structure light patterns nor has a consistent refractive index of the medium which leads to missing depth readings from the conventional depth camera as shown in Figure~\ref{fig:trans_challenge}. As a result, it remains effectively difficult for robots with RGB-D sensors to correctly reconstruct and interact with transparent objects in household settings.

Most existing works~\cite{qian20163d,wu2018full,lyu2020differentiable,li2020through,xu2022hybrid,zhou2019glassloc} on table-top transparent object reconstruction are performed under the lab environment with specialized hardware setup. These approaches have strict prerequisites: known material index of refraction, one or two times of light refraction through an isolated solid object, external light sources, or a background with coded patterns. Apart from those prerequisites, These approaches often acquire object silhouettes from multiple views, perform a space carving/visual hull to initialize a 3D convex shape, and then use the consistency between position and normal to refine the 3D shape further. 

\begin{figure}
    \centering
    \includegraphics[width=\columnwidth]{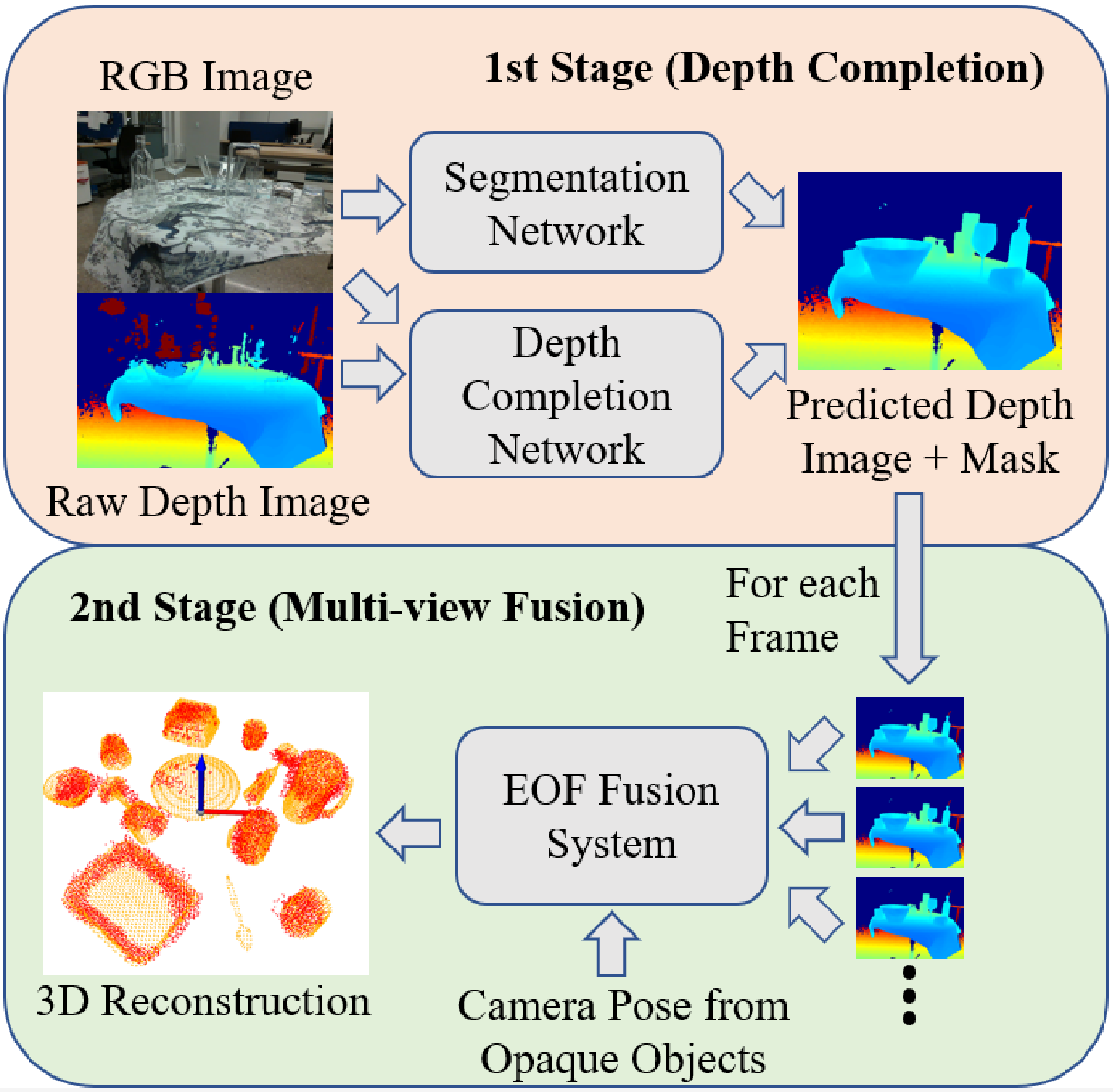}
    \caption{An overview of \trecs transparent object 3D reconstruction pipeline. The first stage predicts depth and segmentation mask from single viewpoints. The second stage generates 2D correspondences and refines through the epipolar-line-guided optical flow for transparent surfaces. The finalized 2D correspondence and camera pose from opaque objects will be jointly optimized through Bundle Adjustment to output the reconstructed point cloud.}
    \label{fig:pipeline}
\end{figure}

By contrast, in this paper, we build a tabletop transparent object scene reconstruction pipeline (Figure~\ref{fig:pipeline}) using a RGB-D camera without any extra external device or strict assumptions on background and lighting conditions. The proposed approach serves as an upstream scene reconstruction module to provide reliable point clouds for downstream robotics reasoning and action over transparent objects. 


More specifically, we propose Monocular Depth Prior-based Epipolar-Guided Optical Flow (\trecsend) as a two-stage tabletop transparent scene reconstruction approach using a sequence of RGB-D images as input. 
In the first stage, off-the-shelf single-view depth completion networks and a segmentation network are employed to acquire transparent object masks and approximate depth predictions. However, due to errors in depth predictions, inconsistencies across views are present at this stage, leading to a noisy reconstruction of the scene.
Addressing this, the second stage formalizes the problem as a Bundle Adjustment (BA) optimization process, focusing on the identification of reliable 2D correspondences on transparent surfaces within the RGB domain. Recognizing the inherent challenges in feature matching on transparent surfaces due to intricate light interactions, we introduce Epipolar-guided Optical Flow (\eofend). \eof integrates two components: (1) Boundary-inspired 2D landmarks generation, which judiciously selects features within transparent regions by capitalizing on boundary information, and (2) Epipolar-guided optical flow correspondence estimation, which monitors the selected landmarks to establish, correct, and refine correspondences utilizing epipolar constraints. This synergistic pairing is pivotal in securing reliable feature matches, which are indispensable for the precise reconstruction of transparent objects.

We assessed \trecs against a state-of-the-art end-to-end transparent object reconstruction method, two state-of-the-art depth completion networks (utilized as our first stage), a general-purpose reconstruction method, and conducted two ablation studies. The comparisons underscored the advantages of our \eof approach. In terms of overall 3D reconstruction accuracy and completeness metrics, \trecs  surpassed these methods, underlining its effectiveness in reconstructing transparent objects.


%% file: 2_relatedwork.tex
\subsection{Transparent Surface Reconstruction}

Reconstruction of transparent surfaces or objects is a challenging problem. Except for intrusive techniques that require physical contact and might destroy the object, most approaches utilize the light reflection or refraction rules. 
For reflection-based methods, one approach is to estimate the 3D geometry based on the specular highlights from a spotlight~\cite{morris2007reconstructing,yeung2011adequate,he20223d}, structured light~\cite{liu2014frequency} or polarization~\cite{xu2017reconstructing,mingqi2022transparent}. Another line of works developed the `scanning from heating' pipeline that determined the shape through laser surface heating and thermal imaging~\cite{eren2009scanning,landmann2021high}. For refraction-based methods, the shape is solved by analyzing the distortion of background patterns using a time-of-flight camera~\cite{tanaka2016recovering} or light-field camera~\cite{zhou2018plenoptic, wetzstein2011refractive}. Some recent works used the RGB camera to capture multiple views to solve light path triangulation~\cite{kutulakos2008theory} and do environment matting~\cite{zongker1999environment} with the help of a turntable~\cite{wu2018full,lyu2020differentiable} or AprilTags~\cite{xu2022hybrid} to get accurate camera poses. 

Without the dependency on external hardware, Li~\textit{et al.}~\cite{li2020through} designed a physically-based deep rendering network that learned the normal of front and back surfaces and fused features from multiple viewpoints to reconstruct the 3D point cloud. However, this work still requires carefully placed isolated objects and ground truth segmentation.
Zhu~\textit{et al.}~\cite{zhu2021transfusion} built a transparent SLAM system on a mixed transparent-opaque object scene, where they trained a segmentation network on transparency, used the opaque parts to solve camera trajectory, and reconstructed the transparent objects using visual hull space carving. Compared to this work, we used single-view depth completion plus multi-view optimization that could deal with concave shapes.

\subsection{Transparent Scene Depth Estimation}

Transparent objects cause missing or inaccurate values on transparency pixels. Before diving into the depth completion task, object detection or segmentation is a prior task to locate where the transparency lies on the image.
Lai \textit{et al.} and Khaing \textit{et al.}~\cite{khaing2018transparent, lai2015transparent} pioneered on transparent object detection using CNNs.
Xie \textit{et al.}~\cite{xie2020segmenting} developed a deep segmentation model that outperformed other baselines and a large-scale segmentation dataset.
ClearGrasp~\cite{sajjan2020clear} tackled the problem of depth completion for transparent object scene. They employed depth completion on robotic grasping tasks, where they trained three DeepLabv3+~\cite{chen2018encoder} models to perform image segmentation, surface normal estimation, and boundary segmentation.
Recently, various different approaches have been explored to improve the speed and accuracy of the depth completion task, including implicit functions with ray-voxel pairs~\cite{zhu2021rgb}, neural radiance field for rendering~\cite{ichnowski2021dex}, combined point cloud and depth estimation~\cite{xu2021seeing}, adversarial learning~\cite{tang2021depthgrasp}, multi-view input~\cite{chang2021ghostpose}, network structure from RGB image completion~\cite{fang2022transcg}, and sim2real transfer~\cite{dai2022dreds}.
Along with the proposed methods, massive datasets, across different sensors and both synthetic and real-world domains, have been collected and made public for various related tasks~\cite{xie2020segmenting,sajjan2020clear,liu2020keypose,zhou2020lit,liu2021stereobj,xu2021seeing,fang2022transcg,chen2022clearpose,dai2022dreds}.

Our system will utilize the advance in deep networks on single-view depth completion and take their output as a prior for further optimization.

%% file: 3_method.tex
Given a sequence of RGB-D input \textit{\textbf{O}} and corresponding camera poses \textit{\textbf{T}} initialized by opaque objects, the objective of \trecs is to infer a point cloud represented by a set of 3D points $\{\textit{\textbf{X}}_i | \text{i=1,...,n}\}$ of the transparent objects.

The two-stage transparent scene reconstruction pipeline is shown in Figure~\ref{fig:pipeline}.
The first stage consists of a single-view depth completion neural network and transparent objects segmentation estimation neural network. The second stage is a transparency-aware 2D landmarks sampling followed by \eof Bundle Adjustment.

\subsection{Single-view Depth Completion and Segmentation}

After evaluating recent advancements in depth completion neural networks~\cite{xu2021seeing,fang2022transcg,sajjan2020clear,zhu2021rgb}, we chose to build the first stage on top of TransCG~\cite{fang2022transcg}, which offers an good balance between processing efficiency and prediction accuracy.
TransCG employs a four-layer U-Net architecture, comprising blocks with intertwined convolutional and fully-connected layers.
The network ingests an RGB image alongside a depth image, and outputs a refined depth image with matching resolution. To regulate both the value and gradient of the predicted depth, the network employs a combined loss $\mathcal{L}$ of depth and surface normal:

\begin{equation}
\begin{split}
\mathcal{L} &= ||\hat{\mathcal{D}} - \mathcal{D}^*||_2 \\
& + \beta (1 - \cos \left< \Delta\hat{\mathcal{D}}_h \times \Delta\hat{\mathcal{D}}_w, \Delta \mathcal{D}^*_h \times \Delta \mathcal{D}^*_w\right>)
\end{split}
\end{equation}
where $\hat{\mathcal{D}}$ and $\mathcal{D}^*$ denote predicted and ground truth depth, and $\Delta\mathcal{D}_w, \Delta\mathcal{D}_h$ denote normalized gradient vectors along depth width-axis and height-axis, respectively. $\beta$ is the weighting factor between two losses. 

As the depth completion network is designed to estimate the depth value at transparent pixels specifically, it is common practice that they require a transparent mask as input. For this purpose, we utilize the widely-adopted Mask R-CNN~\cite{he2017mask} architecture to train a binary segmentation model for transparent objects. Importantly, we fine-tune both TransCG and Mask R-CNN on a dataset specifically consisting of transparent objects to enhance their efficacy. The transparency mask obtained in this stage is also instrumental in the second stage for sampling 2D correspondences.

\subsection{Multi-view Boundary-Inspired and Epipolar-Guided Optical Flow}
Utilizing pre-trained networks that have been fine-tuned on a transparent dataset for single-view depth prediction and transparency segmentation, we are able to directly concatenate the masked predicted depth from multiple views to create an initial 3D shape. However, this direct fusion fails to mitigate the noisy depth predictions and inconsistencies across different views (Figure~\ref{fig:vis_compare} direct concatenation methods). In this section, we introduce the boundary-inspired 2D landmarks generation along with the Epipolar-guided Optical Flow (\eofend) Bundle Adjustment algorithm for transparent object point cloud reconstruction. In this process, we initialize the camera poses from SLAM or dense reconstruction methods over the opaque background, similar to~\cite{chen2022clearpose,zhu2021transfusion}.

\subsubsection{Boundary-Inspired 2D Landmarks Generation}
\label{sec:boundary_landmark}
\begin{figure}[h]
    \centering
    \includegraphics[width=\columnwidth]{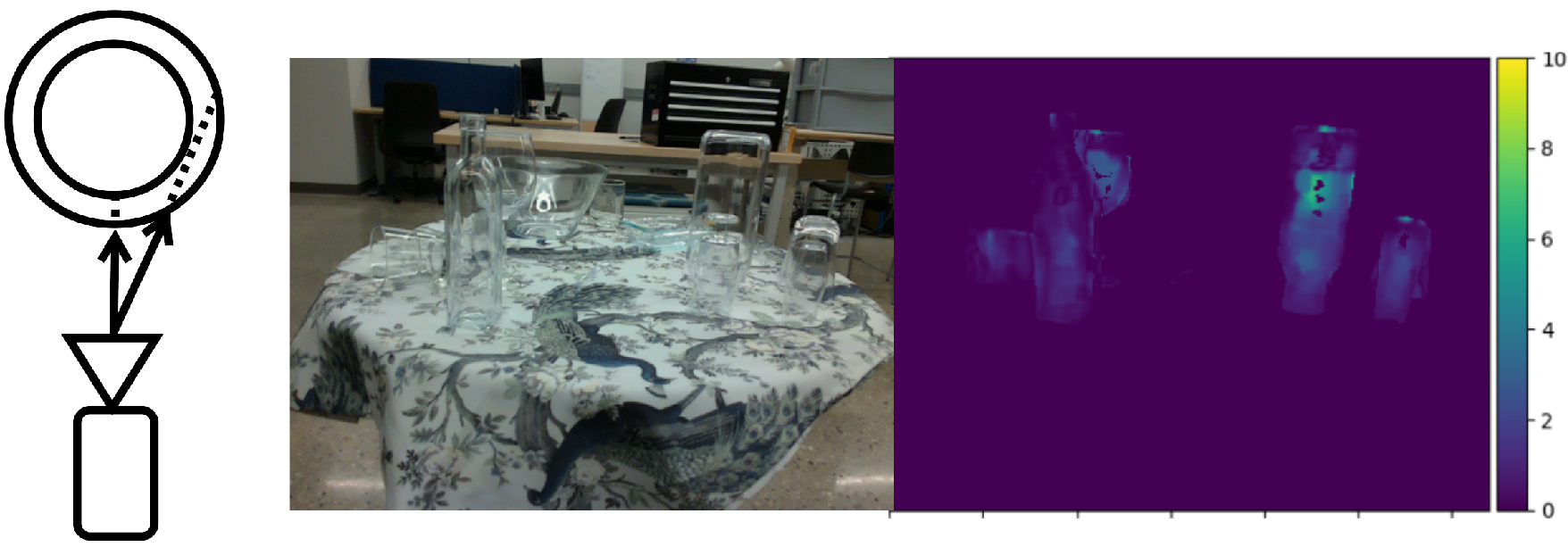}
    \caption{An example that inspires transparent mask boundary-based filtering. (Left) a diagram of camera light rays through a transparent surface, where the effective length is shown as dotted lines. (Middle) an RGB image example of a transparent object scene. (Right) a heatmap corresponding to the RGB image, showing the error of a common dense optical flow estimation between its neighbor frames, measured in pixels.
    }
    \label{fig:boundary_optflow_vis}
\end{figure}

Inspired by how humans perceive transparent objects, we exploit the boundary feature of transparent objects to find reliable visual landmark correspondence between images. As shown in Figure~\ref{fig:boundary_optflow_vis} (Right), the optical flow estimation error is lower in the region near the object boundary while larger in the central areas. 
Our explanation is shown in Figure~\ref{fig:boundary_optflow_vis} (Left). Firstly, for a table-top transparent object, typically a container within the camera field of view, the near-boundary area appears much thicker, which behaves more like opaque surfaces and provides consistent color intensity. Therefore, it provides more reliable correspondence across frames compared to the central area of transparent objects, which corresponds to a relatively thinner surface and is more sensitive to changes in lighting and background. Secondly, the near-boundary area has a distinct appearance compared to the background, which naturally bounds the area in the 2D landmark correspondence search step. 
Based on this observation, we use the pixels near the transparent segmentation mask boundary to generate 2D landmarks. Even though this operation will filter out all 2D landmarks located near the center of the transparent object in a single view, our approach can still obtain enough coverage of objects by incorporating different view angles from a sequence of the RGB-D inputs.

\subsubsection{Epipolar-Guided Optical Flow Correspondence Estimation}
\begin{figure}[h]
    \centering
    \includegraphics[width=\columnwidth]{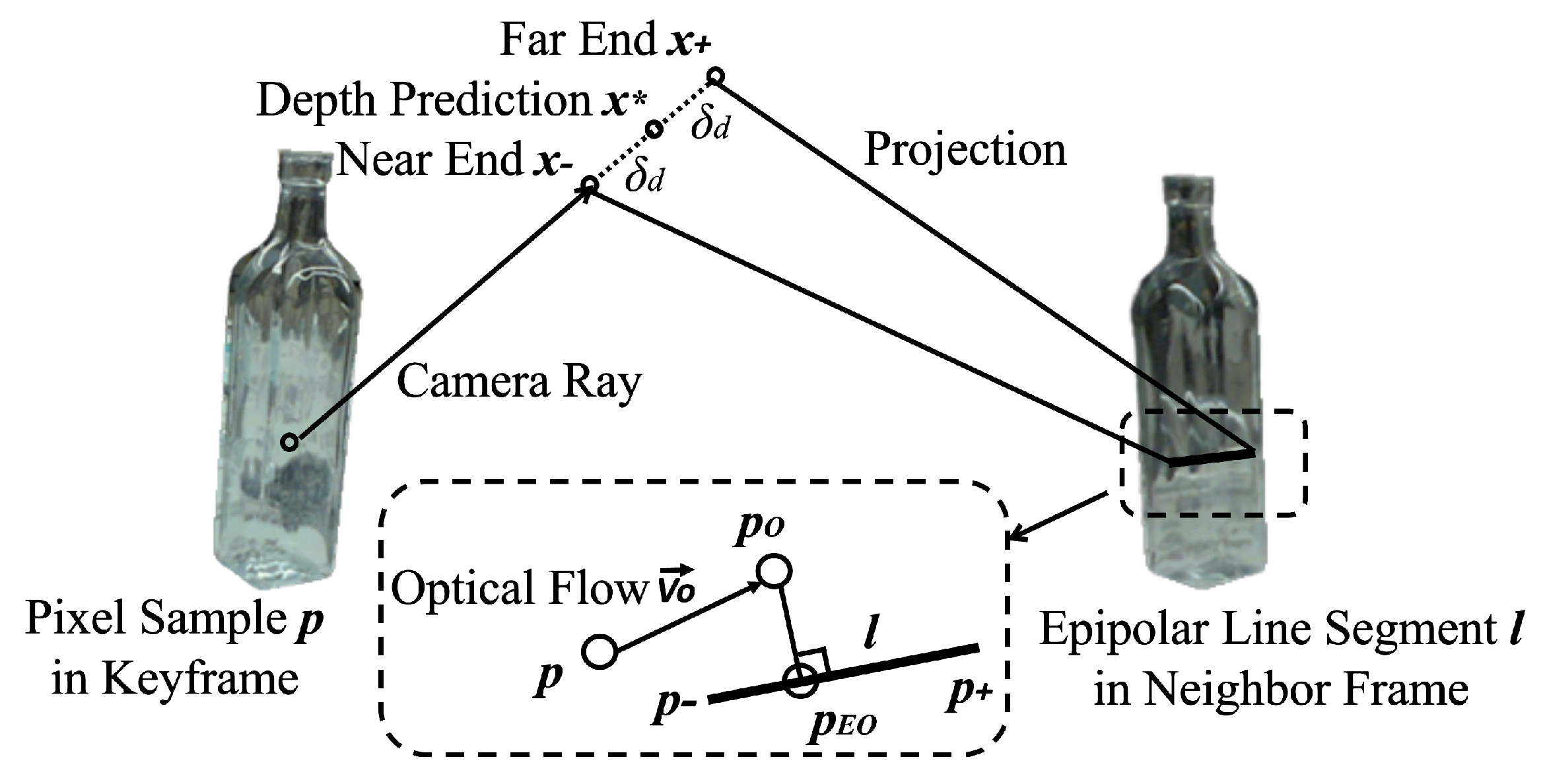}
    \caption{2D correspondence estimation based on epipolar line search and optical flow estimation.}
    \label{fig:epipolar_corr2d}
\end{figure}

We estimate the 2D correspondences within the local epipolar line segment projected from the predicted depth with uncertainty. As shown in Figure~\ref{fig:epipolar_corr2d}, given a 2D pixel sample $\vect{p}$ from one keyframe, its ground truth depth point lies near the 3D points projected using the depth prediction $\vect{x^*}$ on the camera ray. When the depth prediction error is within a certain distance $\delta_d$, the ground truth depth is between the `Near End' $\vect{x^-}$ and `Far End' $\vect{x^+}$. 
Accordingly, the ground truth 2D correspondence lies on the projected epipolar line segment $\vect{l}$ between $\vect{p^-}$ and $\vect{p^+}$ projected by $\vect{x^-}$ and $\vect{x^+}$, respectively, in a neighbor frame when the transform $\vect{T}$ between the keyframe and this frame is known:
\begin{equation}
    \begin{split}
        &\vect{x^\pm} = \left(\hat{\mathcal{D}}\pm\delta_d\right)\vect{K}^{-1}\left[\vect{p}\quad1\right]^T\\
        &\left[\vect{p^\pm}\quad1\right] = s\cdot\vect{K}\left(\vect{R}\vect{x^\pm}+\vect{t}\right)
    \end{split}
\label{eq:proj}
\end{equation}
where $\hat{\mathcal{D}}$, $\vect{K}$ are the single-view depth prediction, camera intrinsic matrix, and $\vect{R}, \vect{t}$ are rotation and translation part of transformation $\vect{T}$, respectively. $s$ is the scale factor (inverse depth) to get the pixel value of $\vect{p^\pm}$.

We then use the dense optical flow estimation $\vect{\Vec{v_0}}$ to get a prediction pixel $$\vect{p_O} = \vect{p} + \vect{\Vec{v_0}}$$ that might not lie on the epipolar line, and take its orthogonal projection towards the line as the final correspondence estimation $\vect{p_{EO}}$. This correspondence will be ignored when the projected point is outside this line segment. The calculation can be formulated as
\begin{equation}
    \vect{p_{EO}} = \vect{p^-} + \dfrac{(\vect{p_O}-\vect{p^-})\cdot\vect{\Vec{n}}}{|\vect{\Vec{n}}|^2}\vect{\Vec{n}}
\end{equation}
where $$\vect{\Vec{n}} = \vect{p^+} - \vect{p^-}$$ is the 2D vector of epipolar line segment.

\subsubsection{Bundle Adjustment Formulation}

We initialize the estimated camera poses from off-the-shelf SLAM systems and formulate a structure-only Bundle Adjustment. The 2D pixel samples $\vect{p}$ in keyframes with valid correspondences $\vect{p_{EO}}$ in neighbor frames are set as the 2D landmarks. Their according 3D landmarks $\vect{x}$ are initialized as $\vect{x^*}$ and optimized by minimizing the reprojection error $\vect{e}$. The optimized 3D points compose the output 3D shape of the \eof module.
\begin{equation}
    \begin{split}
        &\vect{x^*} = \hat{\mathcal{D}}\vect{K}^{-1}\left[\vect{p}\quad1\right]^T\\
        &\vect{e} = \vect{p} - \pi(\vect{T}, \vect{x})
    \end{split}
\end{equation}
where $\pi$ is the projection function same as Equation \eqref{eq:proj}.


%% file: 4_experiment.tex
\subsection{Dataset and Network Training}
\begin{figure}[ht]
    \centering
    \includegraphics[width=\columnwidth]{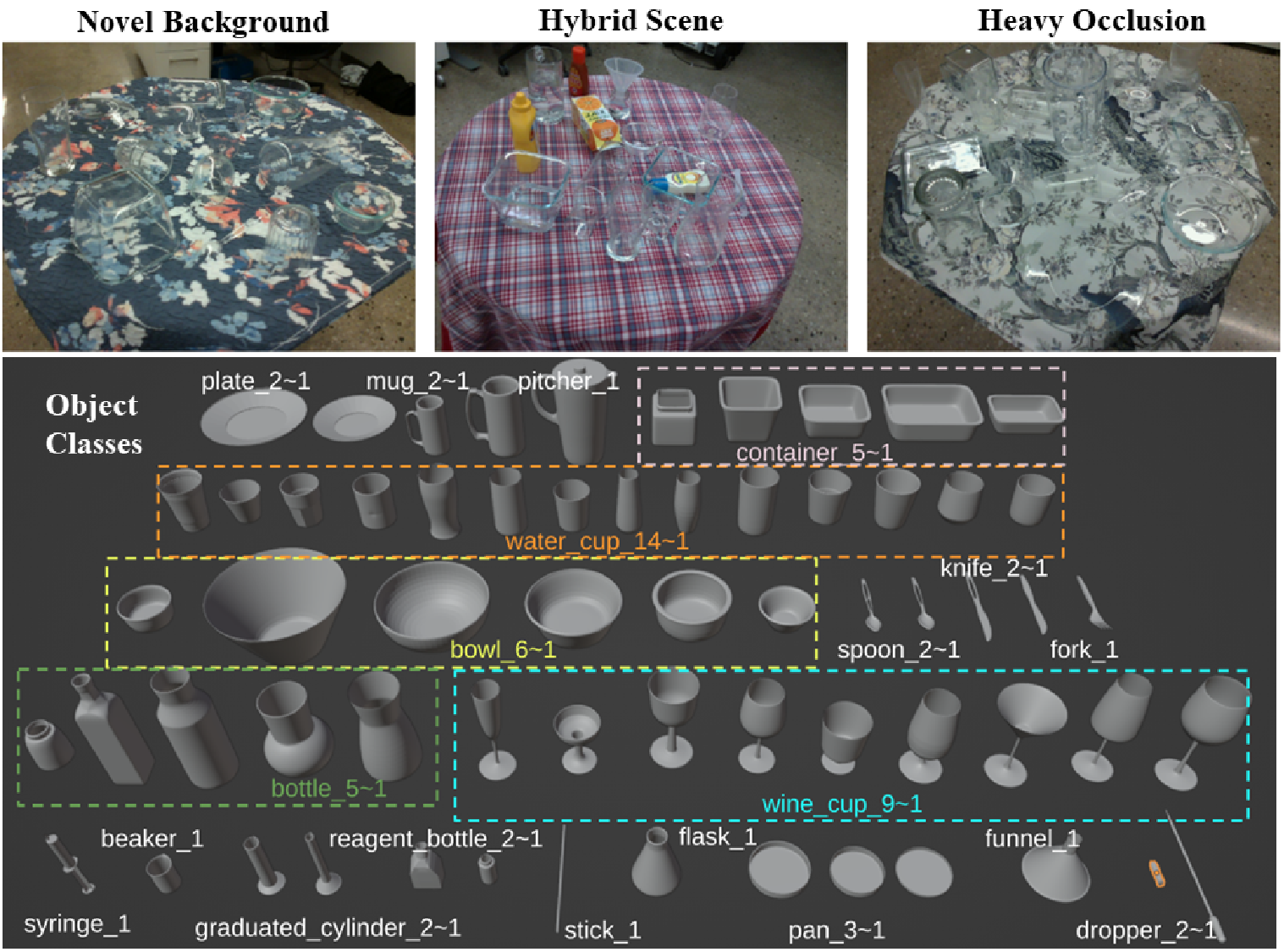}
    \caption{Examples of testing scenes and objects. (Top row) three challenging test scene categories. (Bottom row) object instances that is used in the testing.}
    \label{fig:testsample}
\end{figure}

We evaluate our method using the comprehensive ClearPose dataset~\cite{chen2022clearpose}, a recent collection of continuous RGB-D video frames featuring static transparent object scenes. The training set comprises 25 scenes with 200K images, while the test set spans eight scenes, each containing 1,700 images. The latter encompasses challenges such as novel backgrounds, mixed scenes with opaque objects, and significant occlusions. Figure~\ref{fig:testsample} illustrates examples from the test samples. Notably, even though the training and testing sets include identical object instances, our initial stage is egocentric, employing a depth completion network that remains independent of the object's specific instance.

The network models are implemented in Pytorch. For binary segmentation, we train a Mask R-CNN model\footnote{https://github.com/opipari/ClearPose} on an RTX 2080 SUPER GPU for 5 epochs, with batch size 5 and a learning rate of 0.005 using the SGD optimizer. After training, Mask R-CNN gets an accuracy of Intersection over Union (IoU) as 0.9542 on the training set and 0.8850 on the test set.

TransCG\footnote{https://github.com/Galaxies99/TransCG} is trained on an 8G RTX-3070 GPU for around 300K iterations. $\beta$ is set to 0.001 following the original paper.
We use the $L_2$ loss to train the TransCG network and improve the depth image resizing by using nearest neighbor replacement to maintain the depth discontinuity property near object boundary.
After training, TransCG gets the mean absolute error (MAE) as 0.044m and the root of mean squared error (RMSE) as 0.057m inside the transparent mask on the test set.

\subsection{Evaluation Metrics of 3D Reconstruction}
\label{sec:eval}
We use a combined metrics of Chamfer distance and inlier ratio threshold to evaluate the quality of 3D reconstruction comprehensively. We define 4 values: \textit{Accuracy}, \textit{Completeness}, \textit{Precision}, and \textit{Recall} to evaluate the difference between the predicted 3D point cloud $P_{pred}$ and the ground truth point cloud $P_{gt}$. Chamfer distance is defined as the average of the nearest distance from the points in one point cloud to the points in another point cloud. The inlier ratio is defined as the percentage of points in one point cloud within a certain radius $d$ near the other point cloud. Among the 4 values, \textit{Accuracy} and \textit{Precision} evaluate how close the prediction point cloud $P_{pred}$ is close to the ground truth point cloud, while \textit{Completeness} and \textit{Recall} focus on how much of $P_{gt}$ is covered by $P_{pred}$.

\begin{equation}
\begin{split}
    \text{ChamferDist}(S_1, S_2) &= \dfrac{1}{|S_1|}\sum_{p1 \in S_1} \min_{p2 \in S_2} ||p1 - p2||\\
    \text{Accuracy} &= \text{ChamferDist}(P_{pred}, P_{gt})\\
    \text{Completeness} &= \text{ChamferDist}(P_{gt}, P_{pred})\\
    \text{InlierRatio}_d(S_1, S_2) &= \dfrac{1}{|S_1|}\sum_{p1 \in S_1} \text{I}\left(\min_{p2 \in S_2} ||p1 - p2|| < d\right)\\
    \text{Precision} &= \text{InlierRatio}_d(P_{pred}, P_{gt})\\
    \text{Recall} &= \text{InlierRatio}_d(P_{gt}, P_{pred})\\
\end{split}
\end{equation}

\subsection{Implementation Details and Baselines}

The \eof module is implemented as follows. We build optical flow based on Farneback~\cite{farneback2003two} method. For Bundle Adjustment, we implement it using g2opy\footnote{https://github.com/uoip/g2opy} library with robust Huber loss. 
The 2D landmark points are sampled within the transparent masks on the keyframe RGB images. 
The 3D landmarks in BA are optimized with 30 iterations to get the final 3D point cloud set.
The unoptimized implementation of the 2D landmark generation and refinement took approximately 0.3 seconds per frame to execute on a system equipped with an NVIDIA RTX 2080 GPU and an Intel i7-10875H CPU.


To evaluate the accuracy of the proposed correspondence matching method, we compare it with the SIFT feature detection and matching and optical flow estimation.
To assess the effectiveness of our second stage (\eofend), we integrate it with two state-of-the-art single-frame depth completion approaches: SeeingGlass~\cite{xu2021seeing} and ImplicitDepth~\cite{zhu2021rgb}. For these methods, the baseline directly concatenates the network's outputs across multiple frames.
To evaluate our full pipelines, considering the limited work in multi-view transparent reconstruction, we compare it against one state-of-the-art method on single tabletop transparent object reconstruction: ThroughLookingGlass~\cite{li2020through} and a general-purpose 3D reconstruction method, TSDF, implemented with PyCUDA~\cite{lu2022online}. To make TSDF compatible with transparent objects, we feed the predicted masked depth images from our first-stage network into the TSDF pipeline sequentially and then infer a 3D point cloud surface using the ray marching algorithm\footnote{\scriptsize{https://scikit-image.org/docs/stable/auto\_examples/edges/plot\_marching\_cubes.html}}. We fine-tune the parameters of TSDF fusion to achieve its best performance.
Furthermore, for full pipeline evaluation, we conduct two ablation studies to evaluate how correspondences would influence the final 3D reconstruction output. Specifically, we use SIFT and optical-flow-only methods to refine the correspondences before feeding them into our BA pipeline for 3D reconstruction.

\subsection{Results and Discussions}
\label{sec:exp}
\subsubsection{2D Correspondence Estimation Evaluation}
\label{sec:2d_result}
To validate the reliability of near-boundary pixels for tracking, we measured the 2D correspondence accuracy against the ground truth by projecting its depth onto various frames. Average pixel errors for different frame intervals are presented in Table~\ref{tab:corr2d}. 

\begin{table}[h]
    \centering
    \resizebox{\columnwidth}{!}{%
    \begin{tabular}{ccccccc}\hline
    Frame Interval / Pixel Error & 2 & 4 & 6 & 8 & 10 & 12 \\\hline
    SIFT Feature Only   & 2.380 & 4.192 & 6.184 & 8.364 & 10.782 & 13.418 \\
    Optical Flow Only   & 2.634 & 4.894 & 6.881 & 8.907 & 10.717 & 12.321 \\
    EOF (ours)          & \textbf{0.700} & \textbf{1.254} & \textbf{1.868} & \textbf{2.536} & \textbf{3.124} & \textbf{3.520} \\\hline
    \end{tabular}
    }
    \caption{2D correspondence search accuracy result.}
    \label{tab:corr2d}
\end{table}

In the \textit{SIFT Feature Only} method, we utilized the DoG detector for 2D landmark generation and the SIFT descriptor for matching. The \textit{Optical Flow Only}~\cite{farneback2003two} method initialized the 2D landmarks within the transparency mask and used dense optical flow for matching.


Our proposed method consistently outperforms both baseline approaches across all frame intervals. Traditional techniques tend to diverge rapidly when applied to 2D landmarks on transparent surfaces due to errors from inconsistent RGB observations. In contrast, our method's correction and pruning steps ensure more accurate correspondences.

\subsubsection{EOF Evaluation}
\input{a22_per_scene_with_without_EOBA}
We further evaluate the efficacy of our proposed second stage -- \eof with boundary-inspired sampling on improving overall 3D reconstruction performance with different state-of-the-art single view depth completion networks. The results are shown in Table~\ref{tab:per_scene_results_wo_EOBA}. 
The evaluation is conducted on 8 different scenes, which covers 3 different categories and 14 object classes. We evaluate the performance of different baselines using the metrics described in Section~\ref{sec:eval}.
The results illustrate a notable enhancement
in both accuracy and precision upon the introduction of \eofend, while maintaining comparable recall results.
It is important to note that the direct concatenation method,
despite having a higher point count and thus better completeness, introduces excessive noise which obscures the true shape of the objects.
\input{a3_new_result.tex}

\subsubsection{End-to-end Evaluation}
\input{a1_per_object_evaluation_table.tex}
\input{a2_per_scene_results.tex}

We evaluate the performance of our proposed \trecs pipeline with baseline end-to-end methods using the metrics described in Section~\ref{sec:eval}. 
The results are analyzed on both per object and per scene basis and are presented in Table~\ref{tab:per_obj_results} and Table~\ref{tab:per_scene_results}, respectively.


\trecs outperforms all baseline methods in terms of 3D reconstruction accuracy, precision, completeness and recall at the scene level, as demonstrated by our quantitative results.
Regarding object-level evaluation, our method performs significantly better than all other baselines overall. We also observed that our method performs better for objects with significant height dimension, such as containers, wine cups, and mugs, compared to flat objects like spoons, knives, and forks. 

Figure~\ref{fig:vis_compare} visually demonstrates the performance differences among various 3D reconstruction methods across different scenarios. The first two scenes in each subfigure belong to the \textit{New background} category, while the 3rd and 4th scenes represent \textit{With opaque objects} and \textit{Heavy occlusion} categories. 
ThroughLookingGlass achieves relatively poor results across all four metrics compared to our method. This is because it focuses more on reconstructing a single solid convex object, which does not work well for household objects in a tabletop setup that often have concave shapes and numerous occlusions, leading to missing areas in the reconstruction and very sparse construction. The TSDF method is not specific to transparent 3D construction and produces relatively noisy results, containing many points of the background table and opaque distractor objects, which significantly diminishes its performance. In comparison, our method establishes clear object shapes with fewer noisy points from the background and displays great robustness across heavy occlusions and opaque distractor objects compared to all other baselines.

There are still limitations in our approach. The 2D landmark generation, which uses heuristic filtering based on object boundaries, presumes thicker surfaces near these boundaries. This may not apply to flat surfaces like windows and doors. Also, the approach relies on segmentation masks and demands dense and complete view angle sampling. A confidence map prediction would be preferable to sample with importance and focus on the area with lower error. For 2D correspondence matching, the current optical flow-based implementation hasn't incorporated the recent advances such as deep learning-based methods~\cite{teed2020raft}.
Another point to improve is to consider more than two frames or to involve other modalities, such as surface normal~\cite{li2020through} to refine the correspondence selection.


%% file: a22_per_scene_with_without_EOBA.tex
\begin{table}
\centering
\resizebox{\columnwidth}{!}{%
\begin{tabular}{c@{\qquad}cccc}
  \toprule
  Accuracy (cm)$\downarrow$ & Novel Background & Hybrid Scenes  & Heavy Occlusion  & All Scenes\\
  \midrule
  SeeingGlass~\cite{xu2021seeing} + Direct Concatenation
  & 1.88 & 3.80 & 1.85 & 2.35\\ [0.05cm]
  \textbf{SeeingGlass + EOF}
  & 1.44 & 1.14 & 2.19 & 1.28 \\[0.05cm]
  
  ImplicitDepth~\cite{zhu2021rgb} + Direct Concatenation
  	& 2.10 & 	4.02 & 2.53 & 2.69\\[0.05cm]
   \textbf{ImplicitDepth + EOF}
   & 1.00 & 0.86 & 1.25 & 1.04 \\[0.05cm]

    TransCG + Direct Concatenation
    & 2.63 & 2.06 & 3.96 & 2.45 \\[0.05cm]
  
    \textbf{TransCG + EOF (\trecsend)}
  	& 0.97 & 1.52 & 0.92 & 1.11\\[0.05cm]
  \toprule
  Completeness (cm)$\downarrow$  \\
  \midrule
  SeeingGlass~\cite{xu2021seeing} + Direct Concatenation
  & 0.66 & 0.72 & 0.83 & 0.72\\ [0.05cm]
  \textbf{SeeingGlass + EOF}
  & 1.01 & 1.01 & 0.83 & 1.18 \\[0.05cm]
  
  ImplicitDepth~\cite{zhu2021rgb} + Direct Concatenation
  	& 0.65 & 0.67 & 0.83 & 0.70  \\[0.05cm]
   \textbf{ImplicitDepth + EOF}
 & 0.92 & 1.03 & 0.66 & 0.94 \\[0.05cm]
 
 TransCG + Direct Concatenation
    & 0.51 & 0.48 & 0.50 & 0.59 \\[0.05cm]
    \textbf{TransCG + EOF (\trecsend)}
  	&0.93 & 0.65 & 0.95 & 0.87	 \\[0.05cm]
    \toprule
  Precision (\% $\leq$2cm)$\uparrow$  \\
  \midrule
  SeeingGlass~\cite{xu2021seeing} + Direct Concatenation
  & 64.21 & 44.98 & 62.74 & 59.04\\ [0.05cm]
  \textbf{SeeingGlass + EOF}
  & 78.13 & 82.90 & 69.12 & 77.62 \\[0.05cm]

  ImplicitDepth~\cite{zhu2021rgb} + Direct Concatenation
  	& 62.72 & 43.22	& 53.20 & 55.47\\[0.05cm]
   \textbf{ImplicitDepth + EOF}
    & 86.01 & 89.31 & 80.60 & 84.82 \\[0.05cm]

 TransCG + Direct Concatenation
    & 56.75 & 64.00 & 43.97 & 55.04 \\[0.05cm]
    \textbf{TransCG + EOF (\trecsend)}
  	& 87.38 & 78.05 & 88.41 & 85.09
 \\[0.05cm]
    \toprule
  Recall (\% $\leq$2cm)$\uparrow$  \\
  \midrule

  SeeingGlass~\cite{xu2021seeing} + Direct Concatenation
  & 85.55 & 76.46 & 54.03 & 75.40\\ [0.05cm]
  \textbf{SeeingGlass + EOF}
  & 78.57 & 81.31 & 82.89 & 68.78 \\[0.05cm]
  
  ImplicitDepth~\cite{zhu2021rgb} + Direct Concatenation
  	& 84.55 & 82.08	& 59.21 & 77.61\\[0.05cm]
   \textbf{ImplicitDepth + EOF}
    & 86.33 & 86.76 & 91.67 & 80.12 \\[0.05cm]

   TransCG + Direct Concatenation
   & 95.91 & 96.07 & 99.06 & 92.43 \\[0.05cm]
    \textbf{TransCG + EOF (\trecsend)}
  	& 86.68 & 89.48 & 79.58 & 86.07\\[0.05cm]
   \hline
  \bottomrule
\end{tabular}
}
\caption{Comparison of 3D reconstruction at the scene level, considering results with and without the \eofend. All methods involving \eof are bolded against direct concatenation approaches.
}
\label{tab:per_scene_results_wo_EOBA}
\end{table}

%% file: a3_new_result.tex
\begin{figure*}
    \centering
    \includegraphics[width=\textwidth]{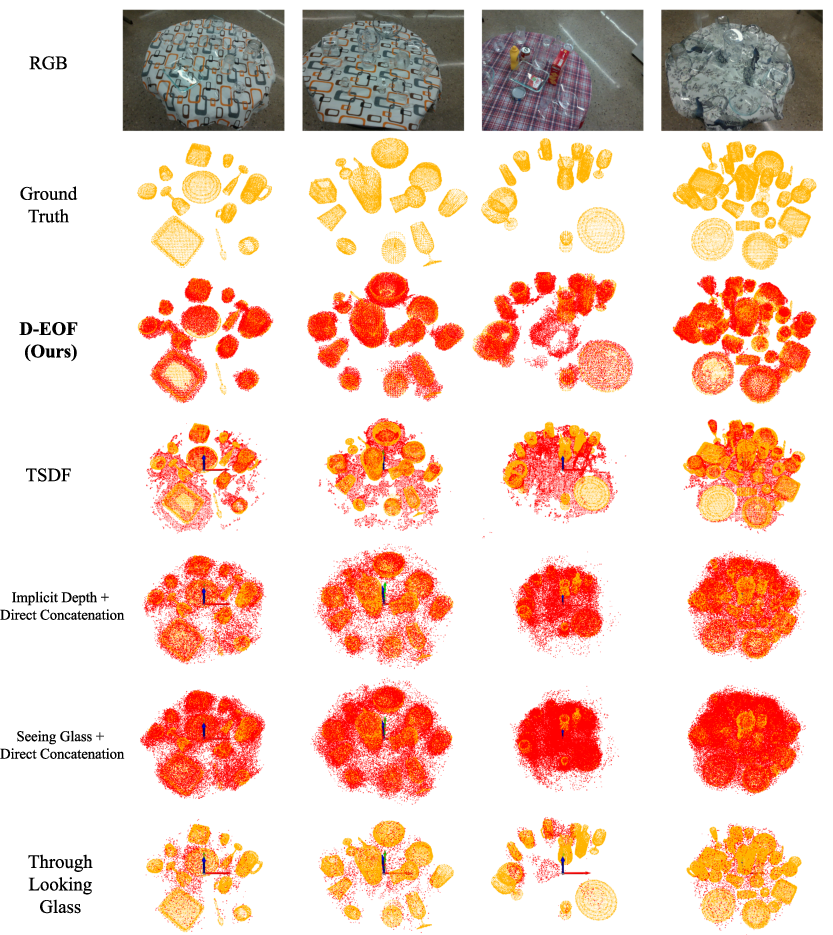}
    \caption{Visualization comparison as 3D point clouds. The orange point clouds represent the ground truth, while the red point clouds depict the predictions of the transparent scenes, corresponding to the RGB images displayed above. It is evident from the visualization that \trecs produces the most consistent and clean results, aligning closely with the ground truth. In contrast, TSDF incorporates numerous points from the background table and opaque distractor objects, substantially reducing its performance. When employing direct concatenation with Implicit Depth and SeeingGlass, the results lack distinction in the object shapes and blanket the entire space. This illustrates that merely concatenating monocular depth priors results in inconsistency and noise across different viewpoints. ThroughLookingGlass generates very sparse points, which can be attributed to its primary focus on single solid convex transparent objects. It struggles to handle cluttered scenes effectively.}
    \label{fig:vis_compare}
\end{figure*}

%% file: a1_per_object_evaluation_table.tex
\begin{table*}
\small
\centering
\resizebox{\columnwidth}{!}{
\begin{tabular}{c@{\qquad}cccccccccccccc}
  \toprule
  Accuracy (cm)$\downarrow$ & Container & Wine Cup & Water Cup & Mug & Plate & Bowl & Bottle & Spoon & Knife & Fork & Pitcher & Funnel & Syringe & Pan \\
  \midrule
  
  ThroughLookingGlass~\cite{li2020through}
  & 1.83 & 1.88 & 1.78 & 1.50 & 2.08 & 1.85 & 1.75 & 2.95 & 2.97 & 2.43 & 1.66 & 1.26 & - & - \\[0.05cm]
  TSDF
  		& 1.27 & 1.62 & 1.48 & 1.09 & 1.49 & 1.60 & 1.47 & 3.51 & 3.43 & 3.59 & \textbf{0.62} & \textbf{1.00} & 2.49 & 2.42 \\[0.05cm]
   SIFT-BA
  		& 1.56 & 1.76 & 1.77 & 1.49 & 1.95 & 1.67 & 1.58 & 2.68 & 2.95 & 3.01 & 1.22 & 1.49 & 3.05 & 2.47 \\[0.05cm]
    OF-BA
  		& 1.64 & 1.63 & 1.68 & 1.44 & 1.88 & 1.74 & 1.57 & 2.36 & 2.70 & 2.78 & 1.11 & 1.84 & 4.07 & 2.97 \\[0.05cm]
    \textbf{\trecs}
  		& \textbf{0.98} & \textbf{1.05} & \textbf{1.04} & \textbf{0.83} & \textbf{0.77} & \textbf{1.04} & \textbf{0.95} & \textbf{0.93} & \textbf{0.86} & \textbf{0.78} & 0.97 & 1.47 & \textbf{0.53} & \textbf{1.40} \\[0.05cm]
  \toprule
  Completeness (cm)$\downarrow$  \\
  \midrule

  
  ThroughLookingGlass~\cite{li2020through}
  & 2.11 & 1.88 & 1.73 & 1.74 & 2.14 & 2.38 & 2.49 & \textbf{1.92} & \textbf{2.99} & 3.21 & 1.65 & 1.27 & - & - \\
  
  TSDF
  		& 1.46 & 2.58 & 1.60 & 1.27 & 2.94 & 1.58 & 1.95 & 2.62 & 4.86 & 2.82 & 0.83 & 1.01 & 1.13 & 4.67 \\[0.05cm]
   SIFT-BA
  		& 1.15 & 1.22 & 1.45 & 1.36 & 1.48 & 1.33 & 1.13 & 2.83 & 3.21 & 2.72 & 0.74 & 1.72 & 3.35 & \textbf{1.54} \\[0.05cm]
    OF-BA
  		& 1.19 & 1.15 & 1.43 & 1.00 & 1.23 & 1.35 & 1.12 & 4.08 & 3.38 & \textbf{1.84} & 0.87 & 0.96 & 4.42 & 2.51 \\[0.05cm]
    \textbf{\trecs}
  		& \textbf{0.68} & \textbf{0.66} & \textbf{0.62} & \textbf{0.48} & \textbf{1.08} & \textbf{0.66} & \textbf{0.56} & 2.53 & 3.10 & 2.14 & \textbf{0.67} & \textbf{0.33} & \textbf{0.96} & 1.79 \\[0.05cm]
    \toprule
  Precision (\% $\leq$2cm)$\uparrow$  \\
  \midrule
  
  ThroughLookingGlass~\cite{li2020through}
  & 59.37 & 58.28 & 61.12 & 72.33 & 50.95 & 62.39 & 61.42 & 47.12 & 67.14 & 48.43 & 63.64 & 76.46 & - & - \\
  
  TSDF
  		& 75.91 & 65.71 & 69.23 & 80.08 & 71.62 & 64.82 & 69.71 & 39.39 & 36.72 & 6.60 & \textbf{91.75} & \textbf{86.13} & 39.53 & 30.89 \\[0.05cm]
   SIFT-BA
  		& 67.66 & 62.51 & 61.41 & 70.14 & 55.28 & 64.58 & 67.66 & 22.92 & 19.39 & 1234 & 79.39 & 70.10 & 4.64 & 35.74 \\[0.05cm]
    OF-BA
  		& 64.28 & 65.40 & 63.94 & 70.90 & 56.39 & 61.95 & 67.52 & 28.02 & 21.16 & 29.83 & 83.80 & 57.73 & 00.00 & 16.77 \\[0.05cm]\textbf{}
    \textbf{\trecs}
  		& \textbf{86.48} & \textbf{84.72} & \textbf{85.19} & \textbf{91.70} & \textbf{94.49} & \textbf{84.68} & \textbf{87.57} & \textbf{87.66} & \textbf{98.62} & \textbf{94.64} & 85.95 & 71.16 & \textbf{100.00} & \textbf{93.46} 
 \\[0.05cm]
    \toprule
  Recall (\% $\leq$2cm)$\uparrow$  \\
  \midrule
  
  ThroughLookingGlass~\cite{li2020through}
  & 19.89 & 27.67 & 30.06 & 29.56 & 18.90 & 19.58 & 28.76 & 9.05 & 04.62 & 17.38 & 15.67 & 65.78 & - & - \\
  
  TSDF
  		& 63.08 & 54.94 & 60.95 & 72.46 & 33.33 & 56.61 & 55.61 & 10.20 & 5.83 & 3.46 & \textbf{86.54} & 81.15 & 59.95 & 2.30 \\[0.05cm]
   SIFT-BA
  		& 66.19 & 66.59 & 58.07 & 59.36 & 67.07 & 63.25 & 68.24 & 30.69 & 26.16 & 11.18 & 80.48 & 32.66 & 00.00 & \textbf{45.33} \\[0.05cm]
    OF-BA
  		& 68.52 & 64.29 & 58.24 & 73.84 & 69.35 & 60.46 & 67.28 & 30.02 & 27.20 & 23.35 & 75.42 & 76.38 & 00.00 & 10.07 \\[0.05cm]
    \textbf{\trecs}
  		& \textbf{87.35} & \textbf{88.45} & \textbf{90.17} & \textbf{94.42} & \textbf{70.48} & \textbf{89.81} & \textbf{91.28} & \textbf{33.95} & \textbf{50.00} & \textbf{25.10} & 86.42 & \textbf{100.0} & \textbf{81.23} & 23.90 \\[0.05cm]
   \hline
  \bottomrule
\end{tabular}
}
\caption{Object-level comparison of 3D reconstruction results across different end-to-end methods.}
\label{tab:per_obj_results}
\end{table*}

%% file: a2_per_scene_results.tex
\begin{table}
\small
\centering
\resizebox{\columnwidth}{!}{%
\begin{tabular}{c@{\qquad}cccc}
  \toprule
  Accuracy (cm)$\downarrow$ & Novel Background & Hybrid Scenes  & Heavy Occlusion  & All Scenes\\
  \midrule
  
  ThroughLookingGlass~\cite{li2020through}
  & 2.38 & 2.75 & 2.75  & 2.56\\ [0.05cm]
  
  TSDF
  	& 2.95 & 4.42  & 1.12	& 2.86 \\[0.05cm]
   SIFT-BA
  	& 2.15 & 3.06 & 1.43 &1.94	 \\[0.05cm]
    OF-BA
  	& 1.95 & 2.58 & 1.66 & 2.02 \\[0.05cm]
    \textbf{\trecs} 
  	& \textbf{0.97} & \textbf{1.52} & \textbf{0.92}	 & \textbf{1.11}\\[0.05cm]
  \toprule
  Completeness (cm)$\downarrow$  \\
  \midrule
  
  ThroughLookingGlass~\cite{li2020through}
  & 1.76 & 3.19 & 1.93  & 2.16\\ [0.05cm]
  
  TSDF
  	& 1.66 & 2.59 & 1.80 & 1.93	 \\[0.05cm]
   SIFT-BA
  	& 1.54 & 1.68 & 1.15& 1.21 \\[0.05cm]
    OF-BA
  	& 1.54 & 1.23 & 1.32& 1.39	 \\[0.05cm]
    \textbf{\trecs}
  	&\textbf{0.93} & \textbf{0.65} & \textbf{0.95} & \textbf{0.87}	 \\[0.05cm]
    \toprule
  Precision (\% $\leq$2cm)$\uparrow$  \\
  \midrule

  ThroughLookingGlass~\cite{li2020through}
  & 55.33 & 56.37 & 44.96  & 53.00\\ [0.05cm]
  
  TSDF
  	& 51.75 & 36.72 & 80.42 & 55.16\\[0.05cm]
   SIFT-BA
        & 54.57	& 48.75 & 73.31 & 63.63 
  		 \\[0.05cm]
    OF-BA
  	& 58.58 & 59.83 & 66.97	& 61.78 \\[0.05cm]
    \textbf{\trecs}
  	& \textbf{87.38} & \textbf{78.05} & \textbf{88.41} & \textbf{85.09}
 \\[0.05cm]
    \toprule
  Recall (\% $\leq$2cm)$\uparrow$  \\
  \midrule

  
  ThroughLookingGlass~\cite{li2020through}
  & 37.88 & 37.87 & 4.92  & 29.64\\ [0.05cm]
  
  TSDF
        & 57.69 & 43.07 & 59.26	& 54.43 \\[0.05cm]
   SIFT-BA
  	& 59.38	& 54.70 & 69.80 & 68.18\\[0.05cm]
    OF-BA
  	& 57.74	& 67.18 & 62.43 & 62.06\\[0.05cm]
    \textbf{\trecs}
  	& \textbf{86.68} & \textbf{89.48} & \textbf{79.58} & \textbf{86.07}\\[0.05cm]
   \hline
  \bottomrule
\end{tabular}
}
\caption{Scene-level comparison of 3D reconstruction results across different end-to-end methods.}
\label{tab:per_scene_results}
\end{table}

%% file: 5_conclusion.tex

We presented a two-stage framework, \trecsend, for reconstructing transparent objects in cluttered tabletop scenes. Our evaluations demonstrate that the proposed \trecs pipeline yields reliable point clouds, adept at handling the challenges posed by tabletop transparent scenes. For further performance enhancement, integrating learning-based approaches for optical flow estimation and incorporating additional modalities, such as surface normals, are promising directions. Additionally, taking into account instance-level object detection and utilizing shape priors could bring about finer details in reconstructions. This development marks a promising step forward in advancing robotic perception and interaction within environments containing transparent objects.